# Dialogue system with humanoid robot

Koki Inoue[1], Shuichiro Ogake[1], Hayato Kawamura[1], Naoki Igo[2]

*Abstract*— Today, as seen in smart speakers, spoken dialogue technology is rapidly advancing to enable human-like interaction. However, current dialogue systems cannot pay attention not only to the content of speech, but also to the way of speaking and eye contact and facial expressions, while watching the facial expressions of the person with whom one is speaking. Therefore, this study participated in a Japanese competition called the "Dialogue Robot Competition" and attempted to develop a dialogue system that includes control of not only the content of speech but also the robot's facial expressions and gaze in order to realize a humanoid robot that can naturally interact with humans. We also discussed the use of the system in customer service work. The "Dialogue Robot Competition 2022" is a competition to develop a dialogue system that uses a humanoid robot to provide tourist information and compete in terms of the naturalness of the dialogue. For the dialogue system developed, a language model was first created from the Japanese version of Wikipedia using the "gensim" library for natural language processing in Japanese. Next, the "Word Rotator's Distance" method and a method using cosine similarity were used to recognize the content of the visitor's speech. Then, for information on tourist attractions, we used basic information on tourist attractions from "Rurubu DATA" provided by the competition's management, and information on restaurants in the area was obtained by Google Maps API. The result of the preliminary round was 10th place out of 13 teams, and we could not advance to the main competition. The reasons for this are thought to be that my dialogue system did not understand what was said and did not respond correctly to the questions I had prepared, that the types of questions I had prepared in advance were few, that the amount of information was not sufficient, and that the responses to the questions were mechanical and not very desirable.

## I. INTRODUCTION

Spoken dialogue technology is advancing rapidly and is being used in smart speakers and smartphone applications. In fact, smart speakers can be used for simple chats and to control home appliances using IoT. On the other hand, there are challenges in utilizing this technology for customer service tasks. Since actual dialogue uses facial expressions, speech patterns, and eye contact, in addition to the content of speech, current voice-only dialogue systems cannot reproduce everyday dialogue. Similarly, it is difficult to replace voice-only dialogue systems in customer service work. This Dialogue Robot Competition 2022 is an effort to develop a more human-like dialogue system by using humanoid robots that can use facial expressions, speech, and eye contact, just as humans do. Furthermore, since the competition is in the form of a competition, it is possible for professionals from universities, companies, and technical colleges who participate in the competition to improve each other's skills. In this research, we aim to participate in this competition and develop a dialogue system that enables natural dialogue. We will also discuss the use of dialogue systems using humanoid robots in customer service work.

## II. DIALOGUE ROBOT COMPETITION 2022

### A. Overview

The Dialogue Robot Competition is a competition to test the dialogue performance of a dialogue system using the humanoid robot shown in Figure 1 The purpose of the dialogue is for the robot to act as a travel agent and recommend one of two sightseeing destinations chosen by the participant, which the participant then chooses. The venue is as shown in Figure 2, and the humanoid robot and the participant engage in a one-on-one dialogue. The detailed rules are described below.

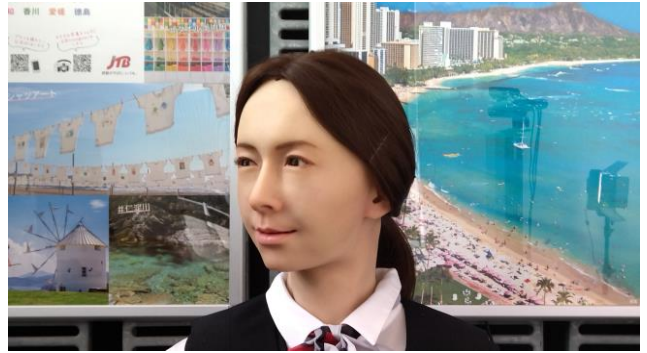

Figure 1. Humanoid robot used in this competition

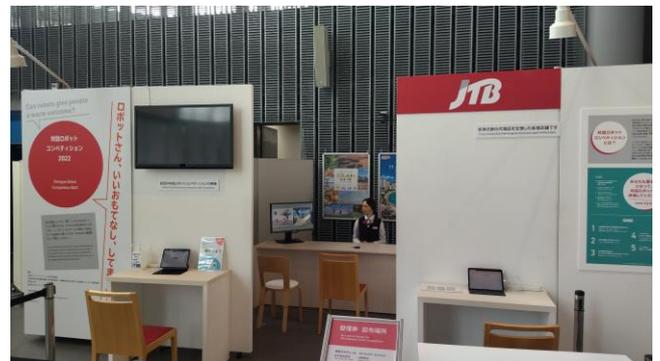

Figure 2. Competition Venues

[1] Department of Production Systems Engineering, National Institute of Technology, Asahikawa College, Asahikawa, Hokkaido, Japan(Tel +81-166-55-8028, E-mail: {p218003 & p218004 & p218007}@edu.asahikawa-nct.ac.jp)

[2] Department of Systems,Control and Information Engineering, National Institute of Technology, Asahikawa College, Asahikawa, Hokkaido, Japan(Tel +81-166-55-8028, E-mail: igo@asahikawa-nct.ac.jp)

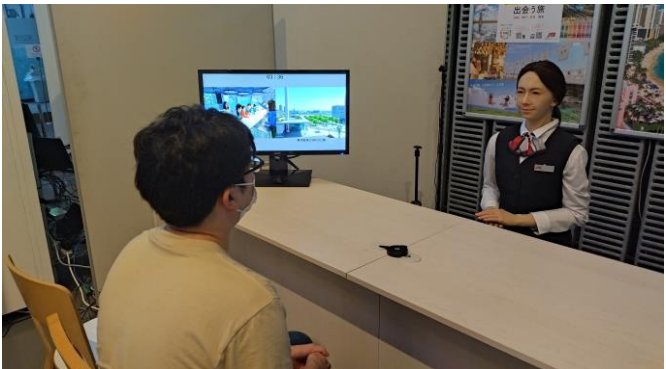
Figure 3. In dialogue

### B. Regulation

The flow of the experience is as follows.
1. Participants select two of the six destinations they would like to visit.
2. participants sit in front of the robot as shown in Figure 3.
3. The participant and the robot interact for a few minutes about the two tourist attractions.
4. After the dialogue, visitors will answer a questionnaire about which tourist destination they would like to visit and about their impressions of the dialogue.

Conduct the dialogue under the following rules.
- The robot and the participant interact one-on-one.
- Dialogue in Japanese.
- The dialogue will take place in a setting where participants will choose two of the six surrounding destinations they would like to visit and discuss which of them they would like to visit.
- The display shows pictures of the two tourist attractions selected by the visitor, and this display cannot be changed.
- The competition is divided into preliminary and final rounds, with the preliminary round taking place at the National Museum of Emerging Science and Innovation Miraikan in Tokyo.

### C. Loaned Software

In the development of the dialogue system, the participants could not directly use the humanoid robot because it was the only robot owned by the management. Therefore, the simulation software and voice recognition software necessary for the development of the dialogue system were loaned to the participants.

*1) AmazonPollyServer*
Software that synthesizes Japanese speech and plays it back on a PC.

*2) CompetitionIDServer*
Software that sends the IDs of the two tourist attractions chosen by the participants and the end time of the dialogue to the client.

*3) Face_recognition*
Software that recognizes participants' faces and sends information on facial expressions to the client

*4) Google Speech Recognition Server*
Software that recognizes participants' voices and sends Japanese text to the client.

*5) MiracleHuman*
Software to simulate upper body movements other than facial expressions of androids

*6) OculusLipSynServer*
Software that simulates mouth movements from voice input to a microphone

*7) JointMapperPlusUltraSuperFace*
Software for simulating the facial expressions of a robot; the robot's facial expressions can be changed by controlling four parameters: valence, arousal, dominance, and realIntention. In addition, by connecting to a client, the facial expressions on the screen can be controlled.

*8) Information on tourist attractions*
Basic information such as names of tourist attractions, hours of operation, and costs were provided by Rurubu DATA.

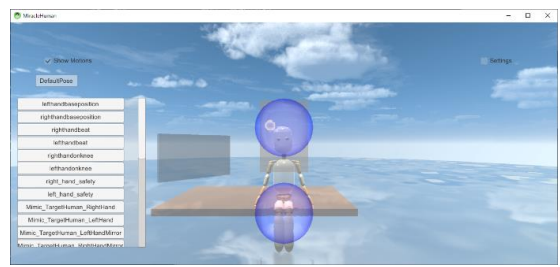
Figure 4. MiracleHuman

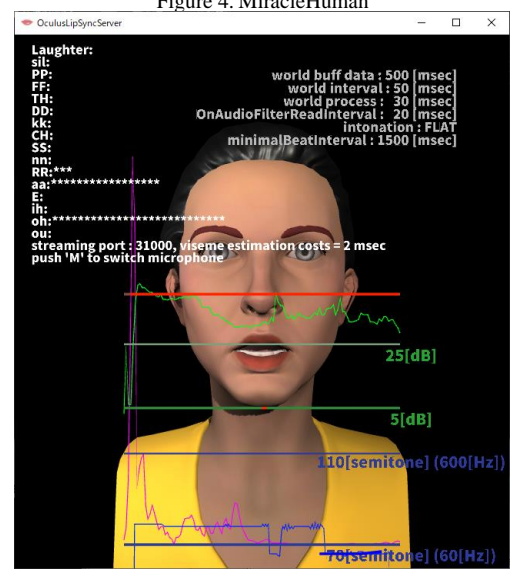
Figure 5. OculusLipSynServer

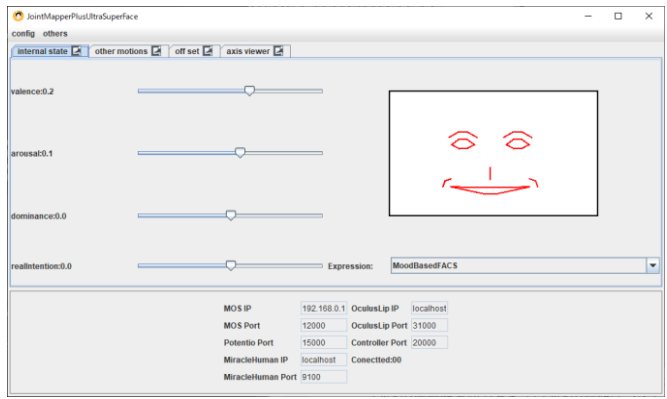

Figure 6. JointMapperPlusUltraSuperFace

TABLE I. EXPRESSION PARAMETERS

|  | valence | arousal | dominance | realIntention |
|---|---|---|---|---|
| smile | 0.3 | 0.2 | 0.1 | 0.0 |
| faint smile | 0.3 | 0.2 | 0.1 | 0.0 |
| surprise | 0.1 | 0.2 | -0.8 | 0.0 |

III. DEVELOPED DIALOGUE SYSTEM

A. Overview

In developing the dialogue system, we first created a Japanese-language model and used natural language processing to analyze visitors' statements in order to understand what they said and what they said about the tourist attractions they were seeking. In doing so, we used basic information on tourist attractions from "Rurubu DATA" provided by a competing operator, and information on nearby restaurants obtained from Google Maps API. The basic flow of the dialogue is shown in Figure 7. After greeting the visitor, the system asks the visitor's name and provides an overview of the two selected sightseeing spots. After that, we ask whether the tourist destination is accessible by car or by train, and in the case of a car, we introduce the "recommended tourist destination" on the grounds that a parking lot is available for cars. In the case of train, I recommend the sightseeing spot because it can be reached by train. After that, the participants answer various questions about sightseeing spots.

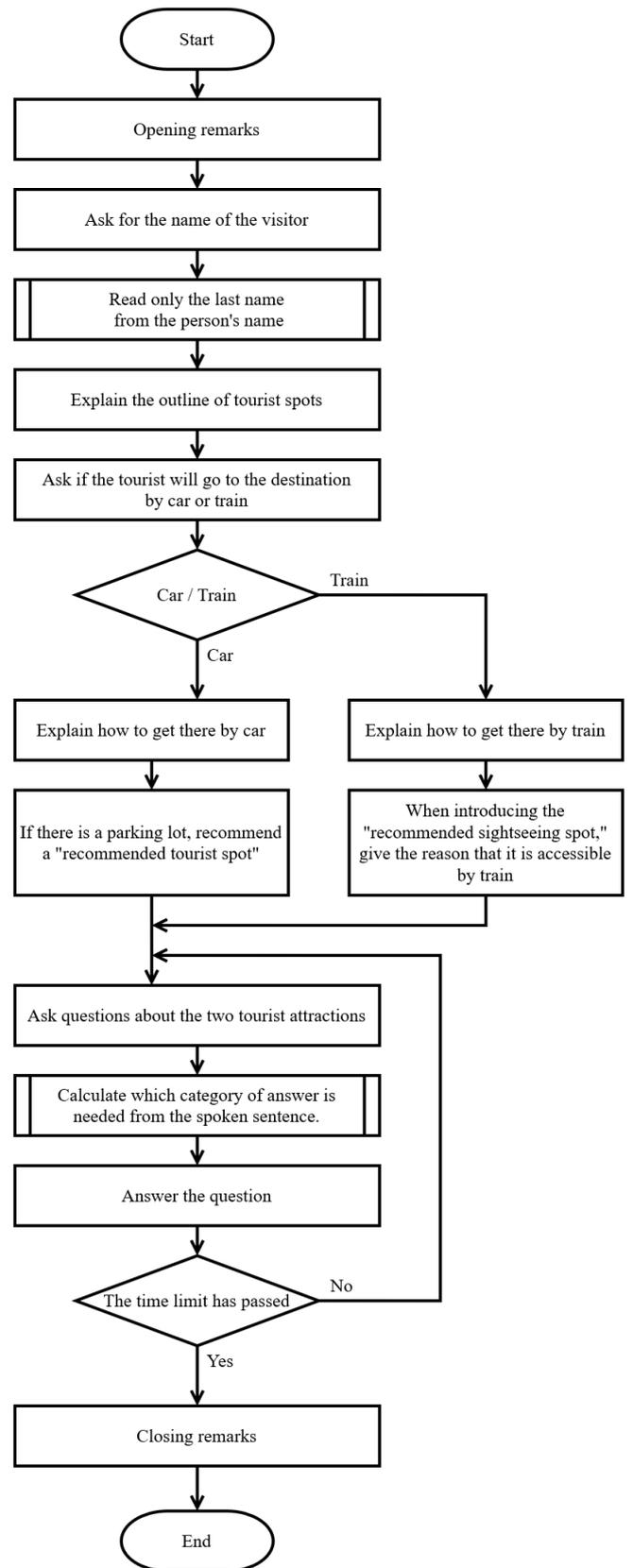

Figure 7. Dialogue flow

*B. Tactics*

In order to receive a high evaluation in the competition, it is necessary to have the "Recommended Tourist Attractions" selected. Therefore, when introducing the two tourist attractions, the "recommended tourist attractions" were explained afterwards to make them more memorable. In addition, in order to increase participant satisfaction, we occasionally called out participants' names.

*C. Japanese Natural Language Processing*

Ubuntu was used as the operating system because it was less trouble than Windows in creating and using a language model, and Visual Studio Code was used as the IDE because it is rich in extensions and allows relatively easy coding, debugging, and source management using Git. The reason for using Visual Studio Code as the IDE is that it is rich in extensions and allows coding, debugging, and source management with Git relatively easily. Python was selected as the programming language because it has a library for natural language processing. The library MeCab is used to perform morphological analysis of Japanese sentences; to use MeCab, a language model is required, and we used the model mecab-ipadic-NEologd. This model is said to be strong for Japanese neologisms, and in fact, it is able to accurately analyze proper nouns such as station names in the dialogue system. gensim is used to understand the content of speech. The use of this library also requires a language model, and we use a language model created from the Japanese version of Wikipedia as of 6/20/2022. By using this language model, we ensure that there are no undefined words in the sentences uttered by visitors.

*D. Recognition of Participants' Statements*

In order to provide information on tourist attractions sought by participants, it is necessary to understand what kind of information they are seeking from the content of their utterances. Therefore, we created categories as shown in Table III for basic tourist attraction information, categorized visitors' utterances into those categories, and had them speak in accordance with those categories. The method for classifying visitors' speech into categories was to create typical questions for each category as shown in Table II calculate the degree of similarity between the questions and the sentences uttered by the participants in the dialogue system, and if there was a category with a high degree of similarity, have the robot utter information on tourist attractions corresponding to that category. If there was a category with a high similarity, the robot was made to utter information on tourist attractions corresponding to that category. Note that the typical question sentences shown in Table II. were actually created in Japanese, and what is shown in Table II is the English translation of the sentences. The calculation of the similarity of those sentences initially utilizes the "Word Rotator's Distance"[1][2] method, which is highly accurate. However, if this similarity does not exceed a threshold value, the method of finding the average cosine similarity of the word vectors in the sentences, which is relatively less accurate but tends to increase the similarity, is used.

TABLE II. SAMPLE CATEGORIES AND CORRESPONDING QUESTION STATEMENTS

| Category | Question statement |
|---|---|
| PriceRemark | How much is the entrance fee? |
| TimeRemark | What are the hours of operation? |
| Parking | Can I park my car there? |

*E. Recognition of Participants' Statements*

For the content of the utterances, the data provided was used for basic tourist information, and information on restaurants in the vicinity was obtained using Google Maps API. For the actual sentences uttered by the robot, the participants' utterances are classified into categories.

## IV. RESULT

*A. Competition Results*

The result of the preliminary round was 10th place out of 13 teams, which did not advance to the final round. The ranking of the results was determined by the total score of the items in Table III. Regarding the evaluation, it was found that this dialogue system was not highly effective in making recommendations, and that the dialogue was not natural. Although the participants were able to understand the content of the dialog to some extent, when they were asked information about one of the sightseeing spots, they responded in an unnatural manner.

TABLE III. RESULT OF PRELIMINARY MEETING QUESTIONNAIRE

| evaluation item | score |
|---|---|
| Choice Satisfaction | 4.3 |
| Sufficiency of information | 4.05 |
| Naturalness of dialogue | 3.85 |
| Adequacy of response | 4.35 |
| Likability of response | 4.1 |
| Dialogue satisfaction | 4.05 |
| Robot reliability | 4.2 |
| referentiality | 4.7 |
| Degree of desire to return | 4.35 |
| Recommended Effects | 8.05 |
| impression total | 37.95 |

*B. Consideration*

Based on the results of the qualifying round, this system is less effective than the other teams in terms of recommendations. Possible reasons for this are as follows. Because the system guided participants to recommended sightseeing spots using only information on how to access the spots and nearby restaurants, it is thought that there was little recommendation information that could be given to the participants. In addition, there were many situations where participants had trouble responding when told in Japanese that everything would be all right. The previous question is considered to be the key, since "It's okay" may include various nuances. If the dialogue system is to be introduced to actual customer service work, it may be easier to talk with the participants by giving them not only information about tourist attractions, but also current trivia.

## V. Conclusion

The purpose of this study was to develop a dialogue system that provides tourist information and to realize a dialogue system using a humanoid robot that can naturally interact with humans for the "Dialogue Robot Competition," in which participants compete to see how natural their dialogue is. In the dialogue system developed, we aimed for natural dialogue by controlling not only the content of speech but also the robot's movements. To understand the participants' speech, a language model was created from the Japanese version of Wikipedia. Next, using the "Word Rotator's Distance" method and cosine similarity, the participants' speech was classified into pre-defined categories, and the participants spoke according to those categories. Basic information on sightseeing spots was obtained from "Rurubu DATA," and information on restaurants in the vicinity was obtained from Google Maps API.

In the preliminary session, we were able to provide an overview of the two selected sightseeing spots, as well as basic information such as access methods and entrance fees, but when asked for information on one or the other, we provided both types of information, making what we wanted to convey unclear. When participants were told in Japanese that "it is OK," it was difficult for them to understand which nuance was implied. The result of the preliminary contest was 10th place out of 13 teams, and the team could not advance to the finals. We believe that the dialogue system developed in this study can turn travel into an easier and more familiar experience and reduce the burden of customer service work.

## Acknowledgment

Thank you to the competition management staff for organizing the Dialogue Competition 2022. I would also like to thank the competition secretariat for their support in participating in this competition. Finally, we would like to thank everyone who helped us in the development of the dialogue system.